\pdfoutput=1
\documentclass[11pt]{article}
\usepackage[final]{acl}
\usepackage{eacl24-tldr-progress-frame}
\graphicspath{{.}}

\begin{document}
\title{\textsc{TL;DR Progress}:\\Multi-faceted Literature Exploration in Text Summarization}

\newcommand{\leipzig}{\textsuperscript{$\dagger$}}
\newcommand{\groningen}{\textsuperscript{$\ddagger$}}
\newcommand{\scads}{\textsuperscript{\S}}

\setlength{\titlebox}{3.6cm}

\author{%
Shahbaz Syed \leipzig\quad
Khalid Al-Khatib \groningen\quad
Martin Potthast \leipzig\scads \\[1.5ex]
\leipzig{}Leipzig University \qquad
\groningen{}University of Groningen \qquad
\scads{}{ScaDS.AI} \\
{\tt\small shahbaz.syed@uni-leipzig.de}}

\date{}

\maketitle

\begin{abstract}
This paper presents {\textsc{TL;DR Progress}}, a new tool for exploring the literature on neural text summarization. It organizes 514~papers based on a comprehensive annotation scheme for text summarization approaches and enables fine-grained, faceted search. Each paper was manually annotated to capture aspects such as evaluation metrics, quality dimensions, learning paradigms, challenges addressed, datasets, and document domains. In addition, a succinct indicative summary is provided for each paper, consisting of automatically extracted contextual factors, issues, and proposed solutions. The tool is available online at {\url{https://www.tldr-progress.de}}, a demo video at {\url{https://youtu.be/uCVRGFvXUj8}}.
\end{abstract}

\section{Introduction}

Research in the field of neural text summarization has evolved rapidly from the introduction of sequence-to-sequence~\cite{sutskever:2014, rush:2015} models to the era of transformers~\cite{vaswani:2017}, greatly improving our ability to produce high-quality summaries in line with human preferences~\cite{huang:2020, goyal:2022}. As a result, the original focus of summarization research on the news domain has broadened to various other domains such as meetings, scientific papers, scripts and opinions.

To keep abreast of current advances, especially researchers new to the field must perform various tasks, including assimilating, organizing, annotating, and reviewing papers across multiple venues. Although search engines tailored to scholarly documents, such as Google Scholar, Semantic Scholar, DBLP, and the ACL anthology, provide access to a vast collection of articles, they merely support the discovery of relevant papers from within a multi-domain collection and do not (strongly) support an in-depth comparative paper analysis.

\begin{figure}[t]
\centering
\includegraphics{conceptual-pipeline-text-summarization}
\caption{Our annotation scheme is based on a summarization literature analysis. Its four components and their respective facets enable a fine-grained unified analysis of relevant papers. The indicative summary is automatically generated.}
\label{fig:conceptual-pipeline-text-summarization}
\end{figure}

This paper introduces {\textsc{TL;DR Progress}}, a literature explorer designed specifically for the text summarization literature. It contributes an intuitive annotation scheme designed to streamline fine-grained, facet-based systematic reviews (Figure~\ref{fig:conceptual-pipeline-text-summarization} and Section~\ref{sec:conceptual-annotation-framework}). To demonstrate the capabilities of our tool, we manually analyze a collection of~514 summarization papers and cover various important aspects for an efficient literature search (Section~\ref{sec:corpus-statistics}). As part of its goal to organize summarization research, {\textsc{TL;DR Progress}} provides an indicative summary for each paper, seamlessly integrating automatically extracted contextual information with manually annotated facets (Section~\ref{sec:indicative-summary}). This includes identifying for what practical purpose a summarization approach is intended, the current challenges associated with summary generation, and a paper's contributions. Our tool also demonstrates the practical application of large language models~(LLMs) in automatic terminology acquisition, involving the extraction of technical terms from papers, including glossary definitions for general concepts and acronym--expansion pairs to improve researchers' recall of specific papers (Section~\ref{sec:generative-retrieval-of-terminology}).%
\footnote{\url{https://github.com/webis-de/eacl24-tldr-progress/}}

\enlargethispage{\baselineskip}
{\textsc{TL;DR Progress}} has a dual function: it provides insights into current research and serves as a basis for future automation. In particular, our literature explorer shows a way forward for future research on large-scale systematic reviews of the NLP literature by extensively leveraging LLMs.

\section{Related Work}

Paper aggregators such as Google Scholar, Semantic Scholar, DBLP, and the ACL Anthology provide access to a large number of papers from different disciplines and, most importantly, facilitate their discovery. However, these platforms lack solid support for in-depth comparative analysis. PersLEARN \cite{shi:2023} introduces a perspective-based approach to exploring scientific literature. It empowers early career researchers to develop their viewpoints by interacting with a prompt-based model. The tool identifies evidence from relevant papers that relate to the given seed perspectives and summarizes them to make new connections. In contrast, \textsc{TL;DR Progress} focuses on summarization and provides fine-grained facets for each paper to enhance understanding of their contributions and content, along with indicative summaries, a feature not present in PersLEARN.

As for annotating papers, Autodive \cite{du:2023} automates the in-place annotation of entities and relationships in PDFs, using external domain-specific NER models. In contrast, our approach includes a domain-specific annotation scheme and manual annotation for quality assurance. In addition, our tool facilitates unsupervised automatic terminology acquisition using LLMs. SciLit \cite{gu:2023} recommends relevant articles based on keywords entered by the user and generates citation sets with extracted highlights. While our tool supports keyword-based lexical search, it is less reliant on user-defined keywords due to its facet-based retrieval system.

Paperswithcode%
\footnote{\url{https://paperswithcode.com/}} is a platform that links papers with their code implementations. It provides an overview of the state-of-the-art in various NLP tasks. \textsc{TL;DR Progress} complements Paperswithcode (for the summarization task) by providing an interactive dashboard presenting relevant statistics (Section~\ref{sec:dashboard-and-figures}) for a comprehensive understanding of the state-of-the-art.

\section{Annotation Scheme}
\label{sec:conceptual-annotation-framework}

\begin{table*}[t]
\centering
\small
\renewcommand{\arraystretch}{1.2}
\setlength{\tabcolsep}{3pt}
\begin{tabular}{@{}lp{0.76\linewidth}@{}}
\toprule
\textbf{Facet} & \textbf{Description / Examples} \\ 
\midrule
\textcolor{webisredA3}{\tt{\textbf{Document representation}}} & \\

\textcolor{webisredA3}{\tt{Input encoding}} &  The paper presents methods to improve the encoding of source documents (e.g., hierarchical/graphical attention, inclusion of discourse structure, etc.) \\

\textcolor{webisredA3}{\tt{Unit relationship}} & The paper investigates methods that explicitly model the relationship between units in the source document, such as words, sentences, or passages.\\

\textcolor{webisredA3}{\tt{Data augmentation}} & The paper introduces methods that use data augmentation techniques, e.g., to extract aspects, to create contrasting examples of robustness, or to overcome data scarcity in low-resource domains. \\

\textcolor{webisredA3}{\tt{External knowledge}} & The paper investigates methods for integrating external knowledge using resources such as knowledge graphs, domain-specific vocabularies, or information from pre-trained language models. \\

\textcolor{webisblueA3}{\tt{\textbf{Model training}}} & \\

\textcolor{webisblueA3}{\tt{Learning Paradigm}} & Supervised, unsupervised, or reinforcement learning. \\

\textcolor{webisblueA3}{\tt{Objective Function}} &  The paper introduces methods that incorporate new objective functions that emphasize diversity, faithfulness, or custom objectives appropriate to the task of summarization. \\

\textcolor{webisblueA3}{\tt{Auxiliary Tasks}} & The paper explores methods such as multi-task learning or pre-training on related tasks (e.g., textual entailment, paraphrasing, gap sentence prediction) to improve the summarization task. \\

\textcolor{webisgreenA3}{\tt{\textbf{Summary generation}}} & \\

\textcolor{webisgreenA3}{\tt{Unit Selection}} & The paper presents methods that explicitly select relevant units, such as words, sentences, or passages, for summarization, addressing the information loss associated with generating fixed-length summaries through techniques such as copying or pointing. \\

\textcolor{webisgreenA3}{\tt{Controlled Generation}} & The paper presents methods that encourage the model to generate summaries with certain attributes (e.g., style, length, tone), for example, by providing additional textual guidance or limiting the model's vocabulary to a specific domain. \\

\textcolor{webisgreenA3}{\tt{Post Processing}} & The paper explores methods for post-processing generated summaries to improve their quality. This includes re-ranking, re-writing or swapping certain text spans to achieve the desired goals. \\

\textcolor{webispurpleA3}{\tt{\textbf{Evaluation}}} & \\

\textcolor{webispurpleA3}{\tt{Domain}} & The domain of the source documents (e.g., opinions, screenplays, papers, etc.) \\

\textcolor{webispurpleA3}{\tt{Dataset}} & The datasets used for training/evaluation (e.g., CNN/DailyMail, XSum, etc.) \\

\textcolor{webispurpleA3}{\tt{Evaluation metric}} & The metrics used for automatic evaluation (e.g., ROUGE, BLEU, etc.) \\

\textcolor{webispurpleA3}{\tt{Human evaluation}} & The summary quality criteria that were evaluated manually (e.g., informativeness, fluency, etc.) \\

\textcolor{mediumgray}{\tt{\textbf{Metadata}}} & \\

\textcolor{mediumgray}{\tt{Paper type}} & A new method, analysis (evaluation), metric, dataset, or theory. \\

\textcolor{mediumgray}{\tt{Venue / Year}} & Venue and year in which the work was published. \\

\textcolor{mediumgray}{\tt{Code / Resources}} & Artifacts relevant to reproduce the paper's contribution. \\
\bottomrule 
\end{tabular}
\caption{Description of the annotation scheme shown in Figure~\ref{fig:conceptual-pipeline-text-summarization}. Pipeline components correspond to the three major components of the scheme, \textcolor{webisredA3}{\tt{Document Representation}}, \textcolor{webisblueA3}{\tt{Model Training}}, and \textcolor{webisgreenA3}{\tt{Summary Generation}}.}
\label{tab:annotation-scheme}
\end{table*}

To create a comprehensive annotation scheme for summarization papers, we performed an in-depth analysis of the recent relevant literature. As shown in Figure~\ref{fig:conceptual-pipeline-text-summarization}, this scheme encapsulates the basic components of a neural summarization architecture, laying the foundation for a fine-grained annotation tailored to individual contributions. Its underlying principle is to categorize contributions according to their main focus, as papers often address one or more components within the summarization pipeline. The annotation scheme distinguishes four components:
\begin{enumerate}
\item
\textbf{Document representation.}
Conversion of a source document into a vector representation to model relationships between document units (words, sentences, paragraphs). This may include input data enrichment with user or style-specific information and model augmentation using external knowledge bases.
\item
\textbf{Model training.}
Training of a model with suitable data under a user-defined objective (or reward) function. This may include using pre-trained models for tasks, such as missing text prediction, paraphrasing, and detecting textual entailment.
\item
\textbf{Summary generation.}
Generation of a summary based on the representation of its source document using a trained model. This may include selecting explicit units for inclusion, restricting the summary to a particular style, conditioning the generation process on certain aspects of the source document, and post-processing steps such as length normalization and redundancy removal.
\item
\textbf{Evaluation.}
Evaluation setup such as the document domains and datasets used for testing, automatic evaluation metrics reported, and the human evaluation criteria for qualitative assessment of the generated summaries. 

\end{enumerate}
These components encompass different facets. For example, document representation includes ``input encoding'', ``unit relationship'', ``data augmentation'', and ``external knowledge''. Definitions for each facet are given in Table~\ref{tab:annotation-scheme}. These facets are not mutually exclusive, i.e., a paper can contribute to several facets simultaneously. For example, a paper may present a novel input encoding scheme that explicitly models unit relationships in the source document. In such cases, we annotate the paper with multiple facets. The annotation scheme also includes metadata for each paper. Overall, our scheme enables a fine-grained retrieval of relevant summarization papers, a feature that is currently not available in other paper aggregators.

\section{Webis Summarization Papers Corpus}
\label{sec:corpus-statistics}

To create {\textsc{TL;DR Progress}}, we compiled a corpus of research papers on neural text summarization, annotated it according to our scheme, and analyzed the distribution of the papers across different dimensions.

\subsection{Corpus Construction}

To collect summarization papers, we conducted a keyword search (``summ'') in the proceedings of the most important venues, including AAAI, AACL, ACL, CHIIR, CIKM, COLING, CONLL, EACL, ECIR, EMNLP, ICLR, IJCAI, IJCNLP, NAACL, NEURIPS, SIGIR, and TACL. The initial collection of 801~papers was refined through a careful review of titles and abstracts to identify papers that were directly relevant to single-document summarization of English texts. These included papers that evaluated or analyzed existing approaches and proposed new metrics, human assessment methodologies, meta-evaluations, datasets, and new model architectures. To extract textual content from the PDFs, we used Science Parse.%
\footnote{\url{https://github.com/allenai/science-parse}}
Papers that could not be automatically extracted or were duplicates were excluded, so that we ended up processing 514~papers. For each of the 514~papers, we performed a thorough manual annotation, focusing on the facets of our annotation scheme. The annotation was performed by one of the authors of this paper. The annotation process was iterative, with the annotator revisiting the previously annotated sections to ensure consistency. Another author reviewed the annotations to ensure their quality.

\subsection{Corpus Statistics}

Table~\ref{table-paper-counts-per-venue} shows the distribution of papers across venues, with EMNLP and ACL emerging as the top venues for summarization research. Among the 514~papers, we observed the following distribution of paper types: 353~dealt with methods, 79~with analysis (including meta-evaluation and quality/model analysis), 73~were corpus-related, 61~focused on metrics and one on theory. The majority of the proposed models were trained using supervised learning~(73\%), compared to unsupervised~(17\%) and reinforcement learning~(10\%). The different paper types were not mutually exclusive, so there were cases where a paper proposed a new dataset and applied methods to it at the same time. In terms of automatic evaluation, the ROUGE metric was used in~71.6\% of papers, highlighting its widespread use for evaluating the quality of generated summaries in the field of single-document summarization of English texts. Only~39.5\% of the papers included some form of manual evaluation. In terms of reproducibility, we found that~58\% of the papers published their code, indicating a slow but growing trend of code availability in this area compared to previous years.

\begin{table}[t]
\small
\centering
\renewcommand{\tabcolsep}{0pt}
\begin{tabular}{@{}lr@{\hspace{2.5em}}lr@{\hspace{2.5em}}lr@{}}
\toprule
  \bf Venue & \bf Count & \bf Venue & \bf Count & \bf Venue & \bf Count \\
\midrule
  EMNLP     &       184 & EACL      &        13 & IJCNLP    &         4 \\
  ACL       &       115 & TACL      &        12 & ICLR      &         2 \\
  NAACL     &        60 & CIKM      &        12 & ECIR      &         2 \\
  COLING    &        34 & AACL      &        11 & NEURIPS   &         2 \\
  AAAI      &        29 & IJCAI     &         9 &           &           \\
  SIGIR     &        17 & CONLL     &         8 &           &           \\
\bottomrule
\end{tabular}
\caption{Number of papers published per venue. Unsurprisingly, EMNLP and ACL are the most popular venues for summarization research.}
\label{table-paper-counts-per-venue}
\end{table}

\begin{table}[!tb]
\centering
\small
\begin{tabular}{@{}p{1\columnwidth}@{}}
\toprule
\textbf{Challenges in Text Summarization} \\
\midrule
Controlled and Tailored Summarization \\
Efficient Encoding of Long Documents \\
Exploiting the Structure of Long Documents \\
Hallucinations in the Generated Summaries \\
Identifying Important Contents from the Document \\
Information Loss / Incoherence in Extractive Summarization \\
Lack of Suitable Training Data \\
Pretraining and Sample Efficiency \\
Robust Evaluation Methods \\
\bottomrule
\end{tabular}
\caption{Manually annotated labels for problem statement clusters extracted from all papers, highlighting the prevalent challenges in text summarization.}
\label{tab:text-summarization-challenges}
\end{table}

\section{Indicative Summaries of Papers}
\label{sec:indicative-summary}

In contrast to informative summaries that aim to replace the entire paper, our tool provides indicative summaries that help users quickly decide if a paper is relevant to their information need. Our indicative summaries are unique in that they encompass an abstractive summary of the paper as well as multiple facets such as datasets, domains, evaluation metrics alongside other information.

\subsection{Beyond Abstract as a Summary}

Traditionally, the paper abstract serves the purpose of an informative summary~\cite{luhn:1958} or an ultra-short abstractive summary~\cite{cachola:2020} that outlines the major contributions. Yet, when dealing with a large collection of documents, these summaries fall short, as they do not enable fine-grained retrieval of relevant papers. Moreover, studies have shown that abstracts can introduce bias and may not offer a comprehensive representation of the paper's contents~\cite{elkiss:2008}.

In contrast to informative summaries, which essentially substitute the source, indicative summaries serve as a roadmap for the contents of the source document~\cite{mani:2001}. They aid readers in deciding whether they want to explore the source document in greater detail. Particularly in the context of literature reviews, indicative summaries provide an exploratory overview of papers, allowing researchers to quickly navigate and comprehend their contributions. {\textsc{TL;DR Progress}} introduces a novel indicative summary that integrates manually annotated facets with automatically extracted contextual information. Motivated by the significance of considering contextual factors in summarization~\cite{jones:1999}, we extract information related to: 
\Ni the purpose of the generated summaries,
\Nii the target audience for the summaries,
\Niii the downstream application of the generated summaries, and
\Niv the problems and corresponding solutions presented in the paper.
Figure~\ref{fig:indicative-summary-paper-example} (Appendix) exemplifies an indicative summary generated by our tool. This summary distinctly outlines all the pertinent information that a reader would need to determine whether they wish to delve into the paper in more detail.

\begin{table}[t]
\small
\centering
\begin{tabular}{@{}p{\linewidth}@{}}
\toprule
\textbf{Context Factors Prompt (GPT3.5)} \\
\midrule
You are a helpful assistant that can read and analyze scientific papers. You are given the following paper: \{\emph{Introduction}\} \newline 
Answer the following three questions: (1) Why are the authors generating the summaries of the documents? (2) Who are they for? (3) How will they be used? You must not include the proposed approach by the authors for generating the summaries. 
\newline
You will output a list of the question-answer pairs where each question is prefixed by the token QUESTION: and each answer is prefixed by the ANSWER: token. Each pair is separated by two lines. \\
\midrule
\textbf{Problems and Solutions Prompt (GPT3.5)} \\
\midrule
\small
You are a helpful assistant that can read and analyze scientific papers. You are given the following paper: \{\emph{Introduction}\} \newline
Can you give me a list of the main problems tackled by the authors and their proposed solutions? In this list, each problem is described followed by a solution proposed by the authors. Each problem starts with the token PROBLEM and each solution starts with the token SOLUTION. \newline 
Here is the list: \\
\bottomrule
\end{tabular}
\caption{Prompts for extracting contextual information from the introduction of a paper. This information is used to compose indicative summaries of papers. The specific instructions for controlling output format may not be required with newer models.}
\label{tab:indicative-summary-prompts}
\end{table}

\subsection{Contextual Information Extraction}

We demonstrate the utilization of LLMs for the task of indicative summarization by extracting the contextual information described above through generative question-answering. To extract this information, we input the introduction section of the paper into the prompt. We devised two prompts corresponding to the \emph{context factors} and \emph{problems and solutions} (see Table~\ref{tab:indicative-summary-prompts}). Each prompt poses specific questions related to the context, necessitating the generation of answers in a specific format using the relevant content from the paper. We employed GPT-3.5 for our experiments.%
\footnote{\url{https://platform.openai.com/docs/models/gpt-3-5}}

We conducted additional analysis of this contextual information to identify the frequently addressed challenges in text summarization. In particular, we employed a soft clustering approach (HDBSCAN~\cite{campello:2013}) on the set of problem statements.%
\footnote{We clustered the contextual embeddings~\cite{reimers:2019} combined with dimensionality reduction using UMAP~\cite{mcinnes:2018}.}
This process yielded 9 clusters, which we manually labeled with their respective challenges, as illustrated in Table~\ref{tab:text-summarization-challenges}.

\section{Automatic Terminology Acquisition}
\label{sec:generative-retrieval-of-terminology}

Scientific terminology plays a vital role in research, requiring researchers to recall papers related to specific concepts or acronyms representing models/metrics. Moreover, previously defined terminology might be directly referenced in subsequent papers without detailed explanation~\cite{ball:2002} or even inaccurately paraphrased, compelling researchers to trace back through multiple papers to find the original definitions. The task of automatic terminology acquisition~\cite{judea:2014} aims to tackle this issue by extracting various concepts defined in a paper along with their definitions. In our exploration of this task, we opted for LLMs instead of supervised methods that necessitate labeled data.

\begin{table}[t]
\small
\centering
\begin{tabular}{@{}p{\linewidth}@{}}
\toprule
\textbf{Glossary Prompt (GPT3.5)} \\
\midrule
You are a scientist who can read and summarize scientific papers. You are given the following paper: \{\emph{Introduction}\}. Your task is to extract a list of key concepts along with correct definitions like a glossary of the paper. Follow the format [\emph{Concept}: \emph{Definition}]. \\
\midrule
\textbf{Acronyms Prompt (GPT3.5)} \\
\midrule
\small
You are a scientist who can read and summarize scientific papers. You are given the following paper: \{\emph{Introduction}\}. Your task is to extract a list of acronyms that the authors use along with correct expansions from the paper. 
\newline 
For example (1)~EDU: Elementary Discourse Unit, (2)~SEHY: Simple Yet Effective Hybrid Model, (3)~PLM: Pretrained Language Model. Exclude acronyms for which no expansion is explicitly provided by the authors. Follow the format [\emph{Acronym}: \emph{Expansion}]. \\
\bottomrule
\end{tabular}
\caption{Prompts for automatic terminology acquisition from the introduction of a paper. We extract glossary as well as acronym-expansion pairs. For the latter, we provide examples of the expected output format.}
\label{tab:terminology-extraction-prompts}
\end{table}

We utilized prompt engineering, leveraging GPT-3.5, with the introduction section of the paper as input for automatic terminology acquisition. We formulated two prompts specifically for extracting \emph{glossary definitions} and \emph{acronym-expansion pairs}.  Examples of extracted glossary terms and acronym-expansion pairs are provided in Table~\ref{tab:examples-of-glossary}. The prompts are shown in Table~\ref{tab:terminology-extraction-prompts}.

\section{Dashboard and Figure Browser}
\label{sec:dashboard-and-figures}

{\textsc{TL;DR Progress}} includes an interactive dashboard that provides real-time visualizations of key statistics gathered from the annotated documents. The dashboard displays:
\Ni the number of papers annotated per year,
\Nii the distribution of publicly released code and resources per year,
\Niii the popular datasets and document domains for training/evaluation,
\Niv the commonly emphasized quality criteria of summary
\Nv the dominant components targeted from the annotation scheme, and
\Nvi the distribution of addressed challenges.

This extensive dashboard delivers a quantitative overview of the text summarization landscape, in line with the detailed facets and additional metadata in our annotation scheme. Key findings from the dashboard include:

\begin{enumerate}
\item Authors consistently practice releasing code for reproducibility and adoption.
\item News (54.2\%) and scholarly documents (13.3\%) dominate as the most studied domains, calling for more diverse investigations.
\item The top three evaluated dimensions for summary quality are informativeness (17\%), fluency (10\%), and coherence (8.1\%).
\item The majority of papers propose new objective functions and input encoding approaches.
\item Predominant challenges include controlled summarization, comprehensive evaluation, insufficient datasets, and risks of hallucinations.
\end{enumerate}

The tool also incorporates a dedicated figure browser (Appendix, Figure~\ref{fig:figure-browser}) hosting \textbf{1524} figures and tables (with captions) linked to their sources. This resource streamlines navigation and serves as a handy reference for researchers exploring standard illustrations depicting model architectures or layouts for presenting evaluation results.%
\footnote{We used \texttt{PDFFigures 2.0}~\cite{clark:2016}.}

\begin{table}[t]
\centering
\small
\renewcommand{\tabcolsep}{2pt}
\renewcommand{\arraystretch}{1.2}
\begin{tabular}{@{}p{0.23\columnwidth}p{0.75\columnwidth}@{}}
\toprule
\textbf{Term}         & \textbf{Definition / Expansion}                                                                                                                                                \\
\midrule
\emph{Glossary} & \\
\midrule
Co-Decoding & An algorithm that takes two review sets as input to compare and contrast the token probability distributions of the models to generate more distinctive summaries \cite{iso:2021}. \\
\raggedright Concept-Pruning & An approach to reduce the number of concepts in a model to find optimal solutions efficiently \cite{boudin:2015}. \\
Drop-Prompt \newline Mechanism & An approach to drop out hallucinated entities from a predicted content plan and to prompt the decoder with the modified plan to generate faithful summaries \cite{narayan:2021}. \\
\raggedright Facet Bias Problem & The problem of centrality-based models tending to select sentences from one facet of a document, rather than important sentences from different facets \cite{liang:2021}. \\
Indegree \newline Centrality & A measure of centrality that assumes a word receiving more relevance score from others is more likely to be important \cite{xu:2020}. \\
\midrule
\emph{Acronyms} & \\
\midrule
ADAQSUM & \uline{Ada}pter-based \uline{q}uery-focused abstractive \uline{sum}\-marization \cite{brazinkas:2022}. \\
COLO  & \uline{Co}ntrastive \uline{l}earning based re-ranking framework for \uline{o}ne-stage summarization \cite{an:2022}.  \\
PLATE   & \uline{P}seudo-labeling with \uline{l}arger \uline{a}ttention \uline{te}mpera\-ture \cite{zhang:2022}. \\
ASGARD  & \uline{A}bstractive \uline{s}ummarization with \uline{g}raph-\uline{a}ug\-mentation and semantic-driven \uline{r}ewar\uline{d} \cite{huang:2020a}. \\
ASAS  & \uline{A}nswer \uline{s}election and \uline{a}bstractive \uline{s}ummariza\-tion \cite{deng:2020}. \\
\bottomrule
\end{tabular}
\caption{Examples of automatically extracted glossary and acronym--expansion pairs from the papers.}
\label{tab:examples-of-glossary}
\end{table}

\section{Evaluation}
\label{sec:evaluation}

We conducted an empirical evaluation of the tool's efficacy in supporting systematic literature reviews for text summarization. The study involved presenting targeted inquiries relevant to beginners in the field and instructing participants to leverage both {\textsc{TL;DR Progress}} and Semantic Scholar for retrieving relevant papers. Additionally, we systematically gathered feedback on the tool's usability and utility for understanding the effectiveness of its features.

\subsection{Purpose-driven User Study}

We conducted a study with five participants (3 PhDs, 2 PostDocs) specializing in natural language processing or information retrieval, but unfamiliar with text summarization research. Their task was to find up to five relevant papers for each of the ten research questions, covering various aspects of summarization research using both {\textsc{TL;DR Progress}} and Semantic Scholar.%
\footnote{\url{https://www.semanticscholar.org/}} 
The following research questions were crafted in reference to the NewSumm Workshop's Call for Papers.%
\footnote{\url{https://newsumm.github.io/2023/}}
\begin{enumerate}
\item How do neural text summarization models address hallucination challenges in abstractive summarization?
\item What are the efficient encoding strategies for handling long documents in neural text summarization?
\item How do neural text summarization models control/tailor the generated summaries to user preferences/aspects/facets?
\item How can pretrained language models be leveraged for improving text summarization?
\item How can additional sources of external knowledge be integrated into the text summarization pipeline?
\item What are the annotation strategies for evaluating hallucination, faithfulness, and factuality in summarization?
\item List at least five corpora that can be used to train scientific document summarization models?
\item List at least five diverse text domains studied in text summarization?
\item What reward functions are proposed to improve summarization via reinforcement learning?
\item What are the various summary quality criteria evaluated via human assessment?
\end{enumerate}

The participants were instructed to evaluate paper relevance using the summaries from the tools, rating them on a scale from 1 (least relevant) to 5 (most relevant). Alongside this, they were requested to share feedback on the usability, the utility of certain features of {\textsc{TL;DR Progress}}, and its strengths and limitations. This evaluation provides both a comparative analysis and a qualitative understanding of the tool's practicality.

\subsection{Results}

Our tool effectively narrowed down the large collection of papers to a set of relevant results. The multi-faceted search, in particular, facilitated quick paper filtering without keyword use. Three out of five participants favored our tool for literature review. However, Semantic Scholar offers a more ``familiar'' search experience and more recent results, albeit requiring extra effort for relevance filtering. Both tools received a score of 4 for the relevance of results. Users also rated the usefulness of features on a scale of 1 (least useful) to 5 (most useful). The advanced search (combining facets) was highly useful (mean score of 4.5), allowing users to easily adapt searches to the research question at hand. This underscores the utility of our annotation scheme (Table~\ref{tab:annotation-scheme}). Indicative summaries and the list of challenges were sufficiently useful (mean score of 3.6) for quickly skimming paper contents and finding papers addressing specific problems, respectively. Results are visualized in the Appendix, Figure~\ref{fig:evaluation-results}.

\subsection{Feedback}

Users found the tool intuitive and easy to use, appreciating the multi-faceted search and indicative summaries. The dashboard was viewed as a useful resource for obtaining a quantitative overview of the text summarization field. Users offered constructive feedback, suggesting incorporating a more sophisticated search mechanism and integrating it with facet-based filtering. They pointed out that searching only by conceptual components was insufficient, as the resulting set of papers was still large and required further filtering. These insights will be considered for future improvements to the tool.

\section{Conclusion}
\label{sec:conclusion}
In summary, \textsc{TL;DR Progress} offers an interactive platform for nuanced exploration of over 500 neural text summarization papers from top venues. Utilizing a tailored annotation scheme, the tool guides users through multifaceted retrieval, provides insightful indicative summaries, outlines challenges, and presents a quantitative overview, easing the entry for newcomers into the field.

\section*{Limitations}
\label{sec:limitations}
The tool leverages LLMs for automated summarization, extracting contextual factors like summary purpose, issues, solutions, and scientific terminology from papers. While we conducted a random accuracy check, a comprehensive assessment of hallucinations or faithfulness in the extracted information was not performed. We anticipate that with more advanced models, such as GPT-4, we can enhance the assurance of quality and structure the extracted content more effectively. Currently confined to summarization, the tool's annotation scheme can be readily extended to other domains, bootstrapped by experts accordingly. However, existing facets like datasets, domains, metrics, qualitative evaluation, and learning paradigms can be directly annotated for new domains. Additionally, a forthcoming feature is the tool's capability to incorporate new papers, automating the annotation process—a feature we plan to implement in future updates to the tool.

\bibliography{eacl24-tldr-progress}

\appendix

\section{Illustrations}
\label{sec:appendix-illustrations}

This section shows the indicative summaries of a paper and the figure browser. For indicative summary, we combined automatically extracted contextual information using prompts.
 
\begin{figure*}[t]
\centering
\includegraphics[height=0.9\textheight]{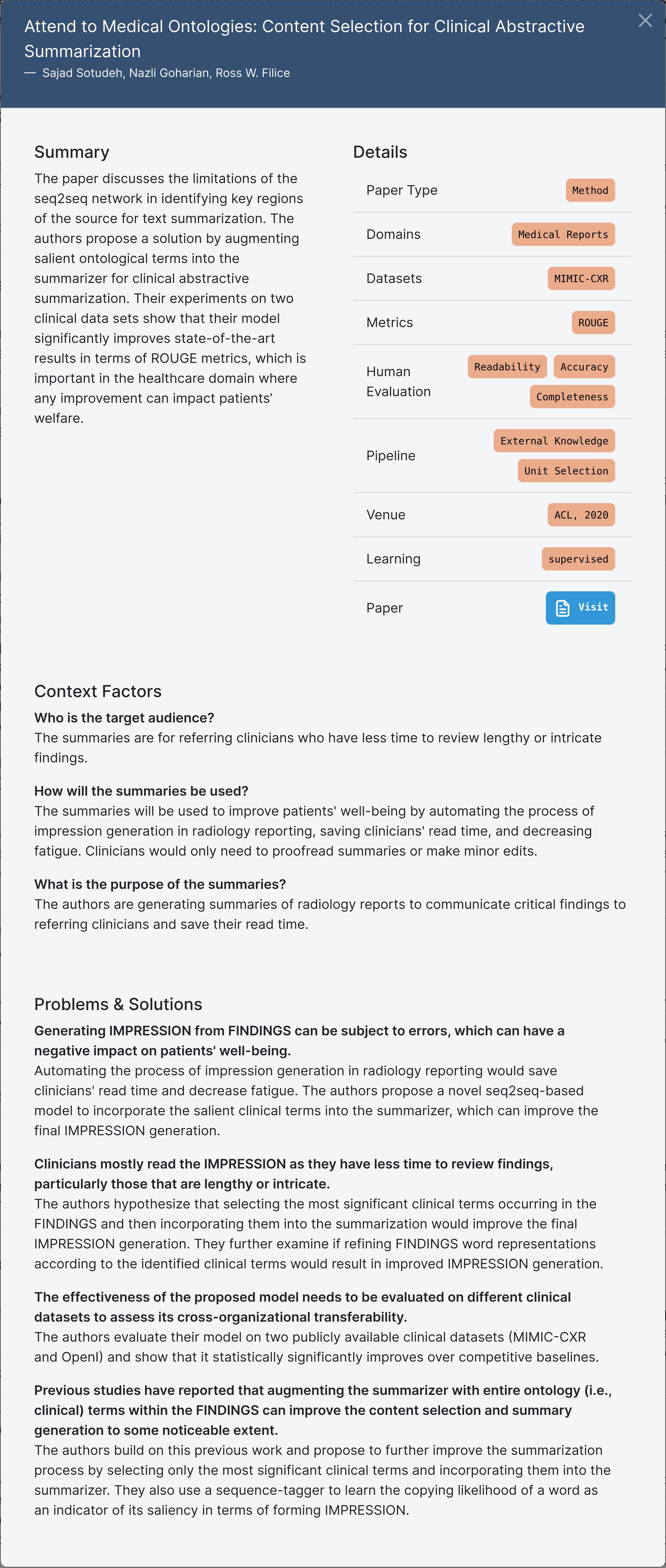}
\caption{Indicative summary of a paper containing
\Ni
an abstractive summary of the introduction,
\Nii
manually annotated metadata attributes (details),
\Niii
purpose of the summary encompassing the target audience, the downstream use, and the purpose, 
\Niv
claims and contributions of the paper.
}
\label{fig:indicative-summary-paper-example}
\end{figure*}


\begin{figure*}[t]
\centering
\includegraphics[height=0.9\textheight]{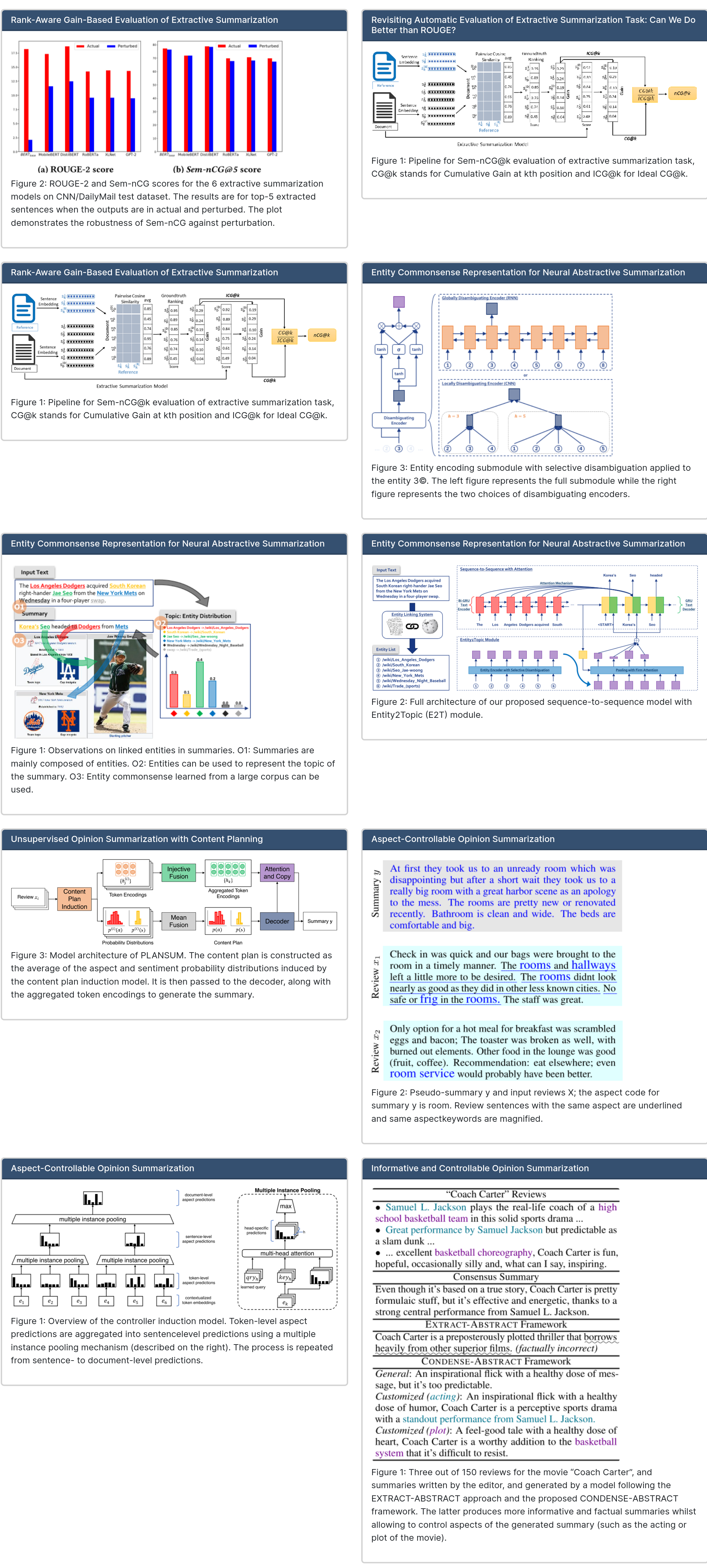}
\caption{An overview of the figure browser which contains all the tables and figures pulled from the papers, accompanied by their captions.}
\label{fig:figure-browser}
\end{figure*}

\begin{figure}[t!]
\small
\begin{subfigure}[t]{\linewidth}
\includegraphics[scale=0.3]{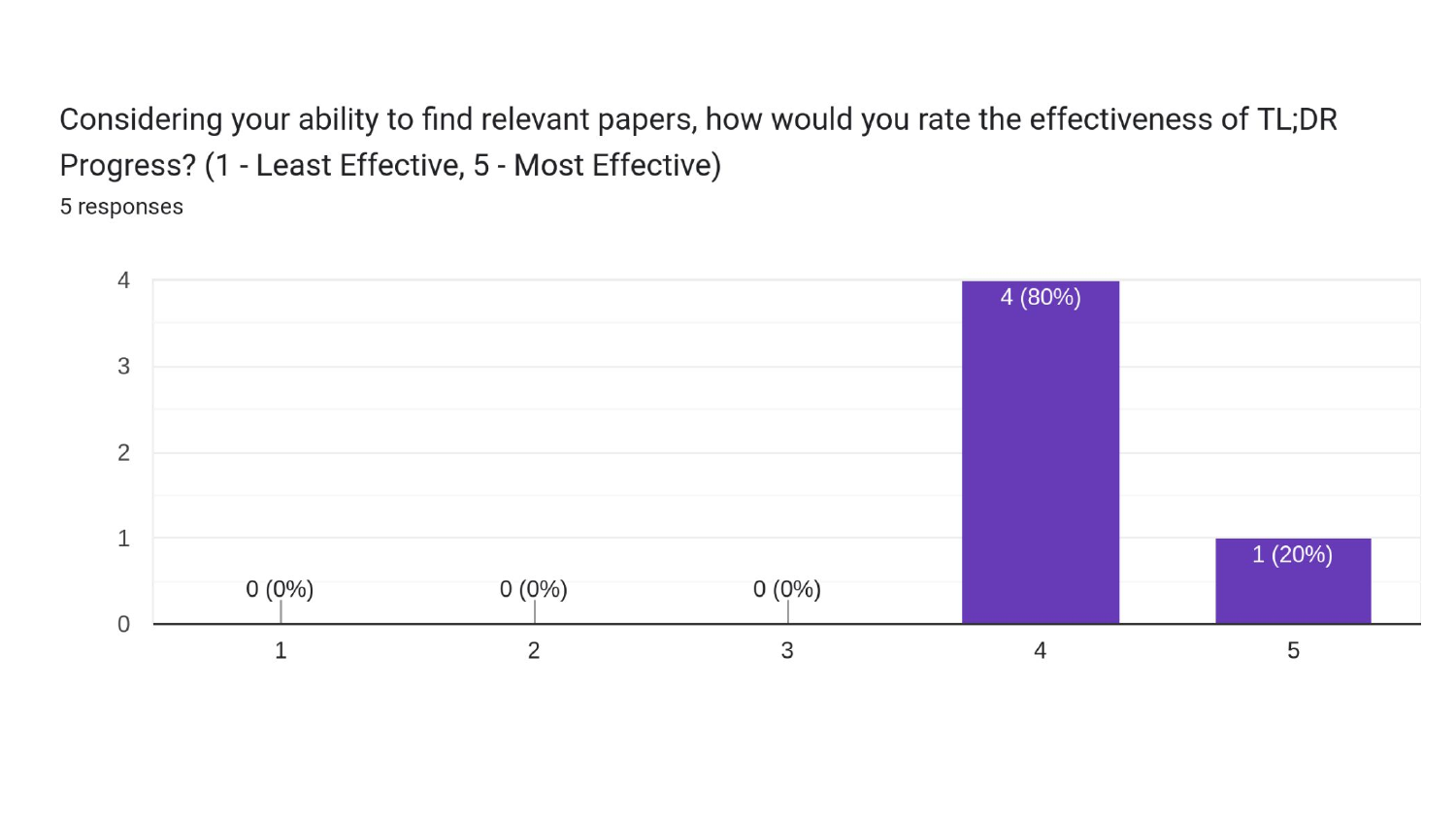}
\caption{Retrieval effectiveness of {\small\textsc{TL;DR Progress}}.}
\label{fig:evaluation-effectiveness-tldr-progress}
\end{subfigure}
\begin{subfigure}[t]{\linewidth}
\includegraphics[scale=0.3]{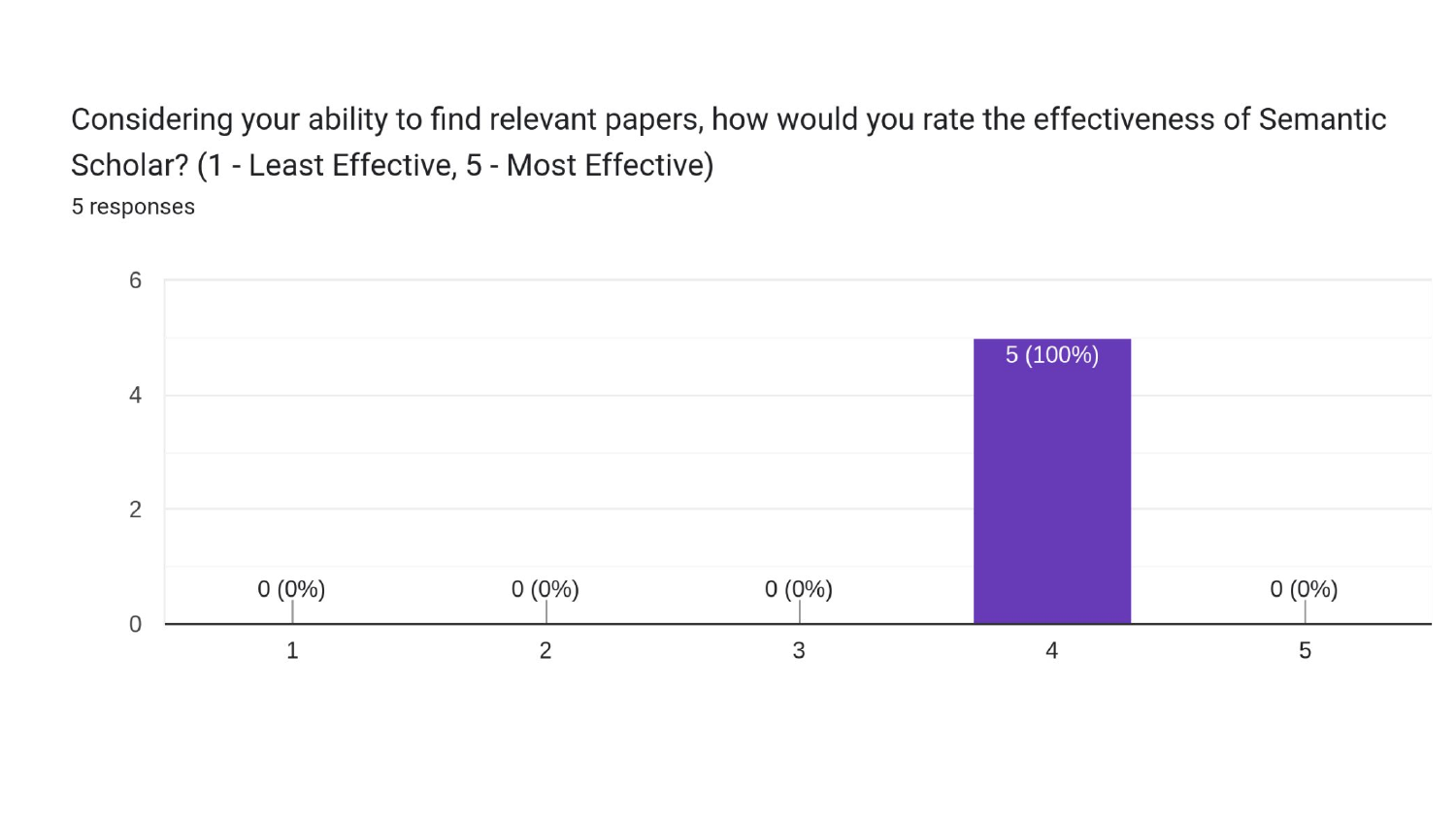}
\caption{Retrieval effectiveness of Semantic Scholar.}
\label{fig:evaluation-effectiveness-semantic-scholar}
\end{subfigure}
\begin{subfigure}[t]{\linewidth}
\includegraphics[scale=0.3]{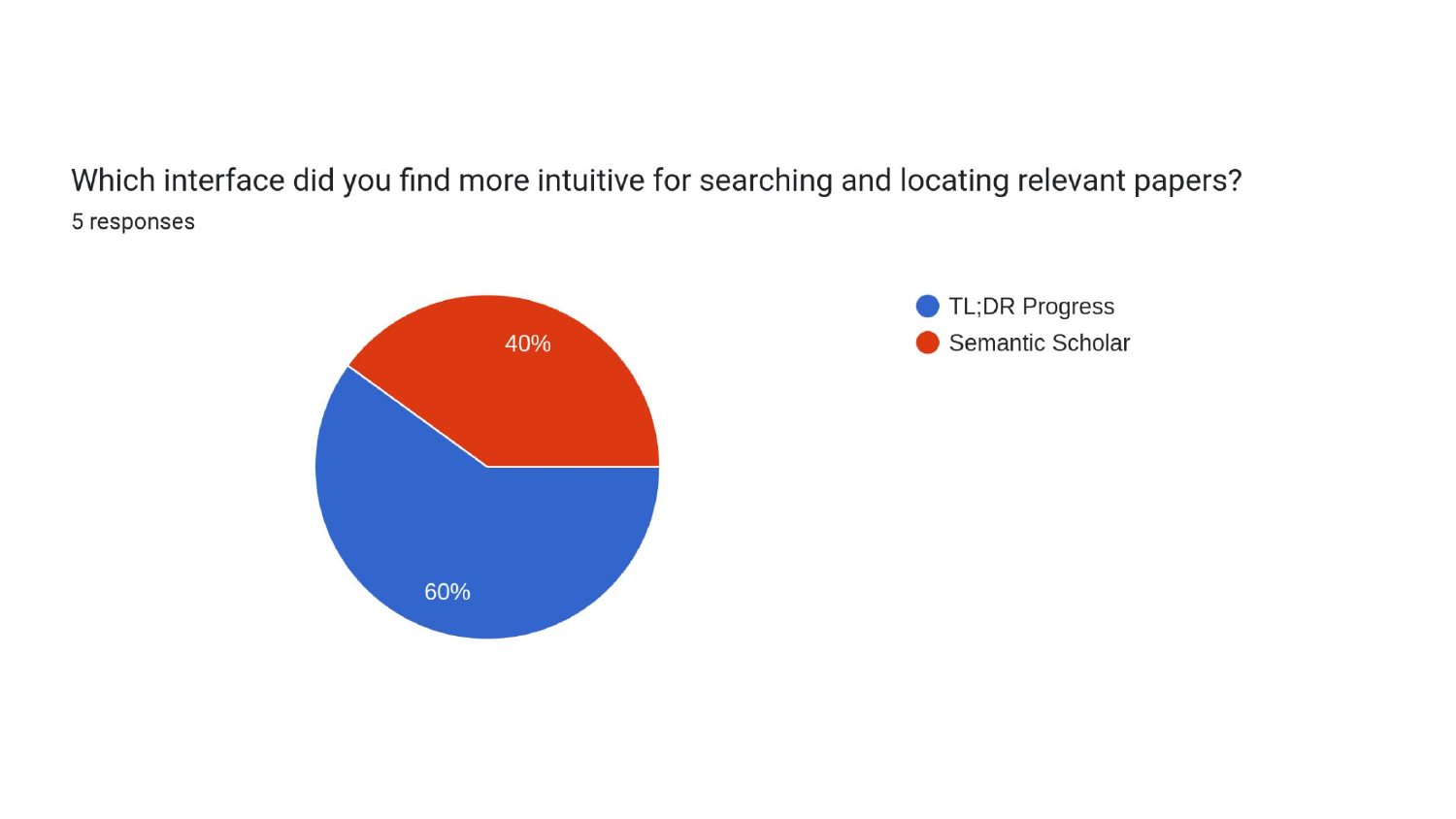}
\caption{Preference of {\small\textsc{TL;DR Progress}} over Semantic Scholar for literature review.}
\label{fig:evaluation-preference}
\end{subfigure}
\begin{subfigure}[t]{\linewidth}
\includegraphics[scale=0.3]{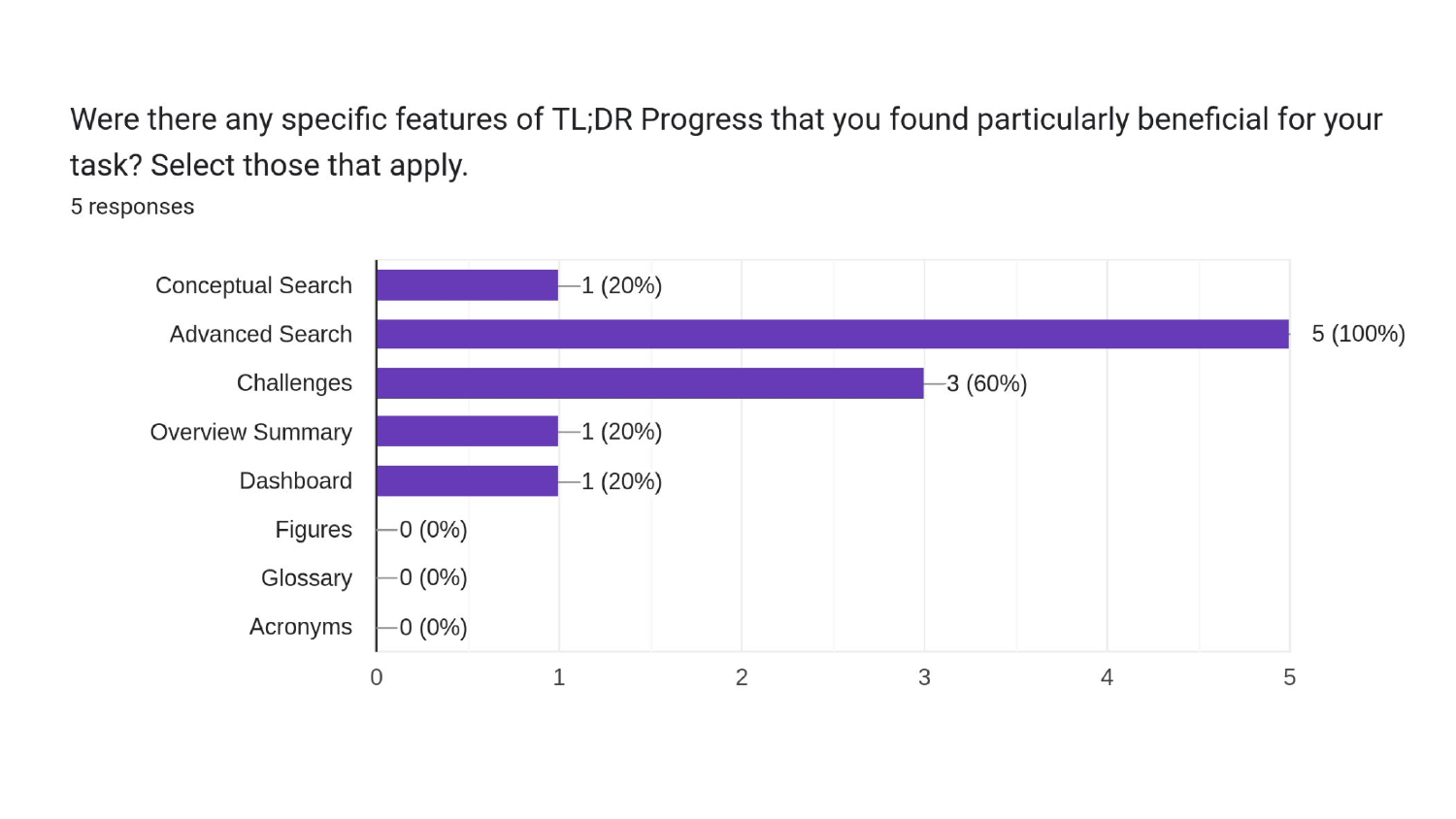}
\caption{Usefulness of features in {\small\textsc{TL;DR Progress}}. Advanced search which allows for combining multiple facets for filtering papers is the most useful feature, followed by the enumerated list of challenges.}
\label{fig:evaluation-usefulness-of-features}
\end{subfigure}
\caption{Evaluation results of the effectiveness and usefulness of {\small\textsc{TL;DR Progress}} compared to Semantic Scholar.}
\label{fig:evaluation-results}
\end{figure}
\end{document}